\documentclass[10pt,twocolumn,letterpaper]{article}

\usepackage{cvpr}

\usepackage{microtype}
\usepackage{graphicx}
\usepackage{subfigure}
\usepackage{booktabs} 
\usepackage{multirow}

\usepackage{times}
\usepackage{epsfig}
\usepackage{graphicx}
\usepackage{amsmath}
\usepackage{amssymb}
\usepackage{overpic}
\usepackage{url}
\usepackage{enumitem}
\usepackage{algorithm}
\usepackage{algorithmic}

\usepackage{xcolor}

\usepackage{hyperref}

\def\x{{\mathbf x}}
\def\z{{\mathbf z}}

\def\u{{\mathbf u}}
\def\v{{\mathbf v}}

\def\M{{\mathbf M}}
\def\m{{\mathbf m}}
\def\0{{\mathbf 0}}

\def\P{{\mathcal P}}

\def\epsx{{\epsilon_{\x}}}
\def\epsw{{\epsilon_{\om}}}
\def\T{{\mathcal{T}}}

\def\dt{{\boldsymbol \delta}}
\def\tt{{\boldsymbol \theta}}
\def\om{{\boldsymbol \omega}}

\def\L{\mathcal{L}}

\def\wrt{\emph{w.r.t.}}
\newcommand{\alg}[1]{\textbf{\texttt{#1}}}
\newcommand{\salg}[1]{{\small\textbf{\texttt{#1}}}}

\cvprfinalcopy 

\ifcvprfinal\fi

\begin{document}

\title{Joint Adversarial Training: Incorporating both Spatial and Pixel Attacks}

\author{Haichao Zhang  \quad  Jianyu Wang \\
	{\fontsize{11}{12} \selectfont \begin{tabular}{c}
			{Baidu Research, \,Sunnyvale USA}
		\end{tabular}} \\
		{\fontsize{9.5}{10}	\texttt{hczhang1@gmail.com\, wjyouch@gmail.com}}\\
	}
\maketitle

\begin{abstract}
Conventional adversarial training methods using attacks that manipulate the pixel value directly and individually, leading to models that are less robust in face of spatial transformation-based attacks.
In this paper, we propose a joint adversarial training method that incorporates both spatial transformation-based and  pixel-value based attacks for improving model robustness.
We introduce a spatial transformation-based attack with an explicit notion of budget and develop an algorithm for  spatial attack generation. 
We further integrate both pixel and spatial attacks into one generation model and show how to leverage the complementary strengths of each other in training for improving the overall model robustness. 
Extensive experimental results on different benchmark datasets compared with state-of-the-art methods verified the effectiveness of the proposed method.
\end{abstract}

\section{Introduction}
While breakthroughs have been made in many fields such as image recognition leveraging deep neural networks, these models could be easily fooled by the so call adversarial examples~\cite{szegedy2013intriguing}.
In terms of the image classification models, an adversarial example
for a given image is a modified version that causes the classifier to produce a label different  from the original one while visually indistinguishable from it.
Previous work mainly focused on improving the model robustness to pixel value perturbations~\cite{FGSM, tramer2017ensemble, madry2017towards, athalye2018obfuscated}.
In contrast, very few work has been done on the model robustness \wrt  spatial transformations with some initial investigation in a few recent works on adversarial attacks~\cite{rot_tran, xiao2018spatially}. 
While it has been shown that certain spatial manipulations of images such as rotation and translation~\cite{rot_tran} or non-rigid  deformation~\cite{xiao2018spatially} can be used to generate adversarial examples for attacking purpose, no practical approach has been developed yet on how to incorporate the spatial domain into the adversarial training framework to further improve the model robustness.
Part of the reason lies in the fact that current works~\cite{rot_tran, xiao2018spatially} are mainly designed for attacking purpose, therefore the cost function and the optimization algorithm therein might not be proper for robust training. For example, \cite{rot_tran} used grid search for the optimization of the transformation parameters which is clearly limited to small parameter space and not scalable. \cite{xiao2018spatially} proposed to generate adversarial examples by smoothly deforming a benign image using a flow field. For this purpose, the cost function used in~\cite{xiao2018spatially} incorporates a smoothness regularization term for the flow field to implicitly encourage the visual similarity. However, in order to get reasonably good solutions, more expensive solvers need to be used for  minimizing the cost function~\cite{xiao2018spatially}. Moreover, the implicit penalty is not directly connected with a quantifiable measure on the strength of the attack, which is important for performing quantitative evaluations of model robustness and benchmarking the performances of different algorithms.
In the following, we will use \emph{pixel attack} to refer to the conventional per-pixel additive attack and use \emph{spatial attack} as a shorthand for the spatial transformation-based attack.

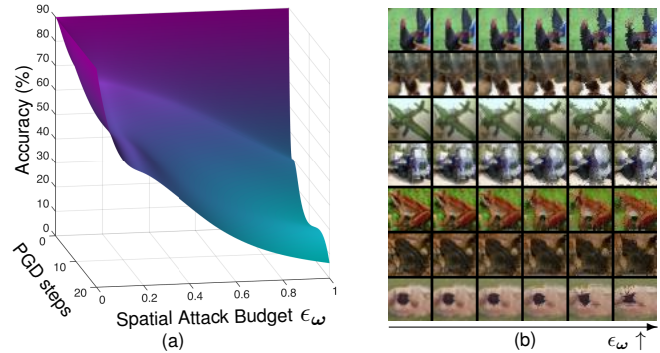
\begin{figure}[t]
	\centering
	\begin{overpic}[viewport = 35 5 550 510, clip, width=1.7in]{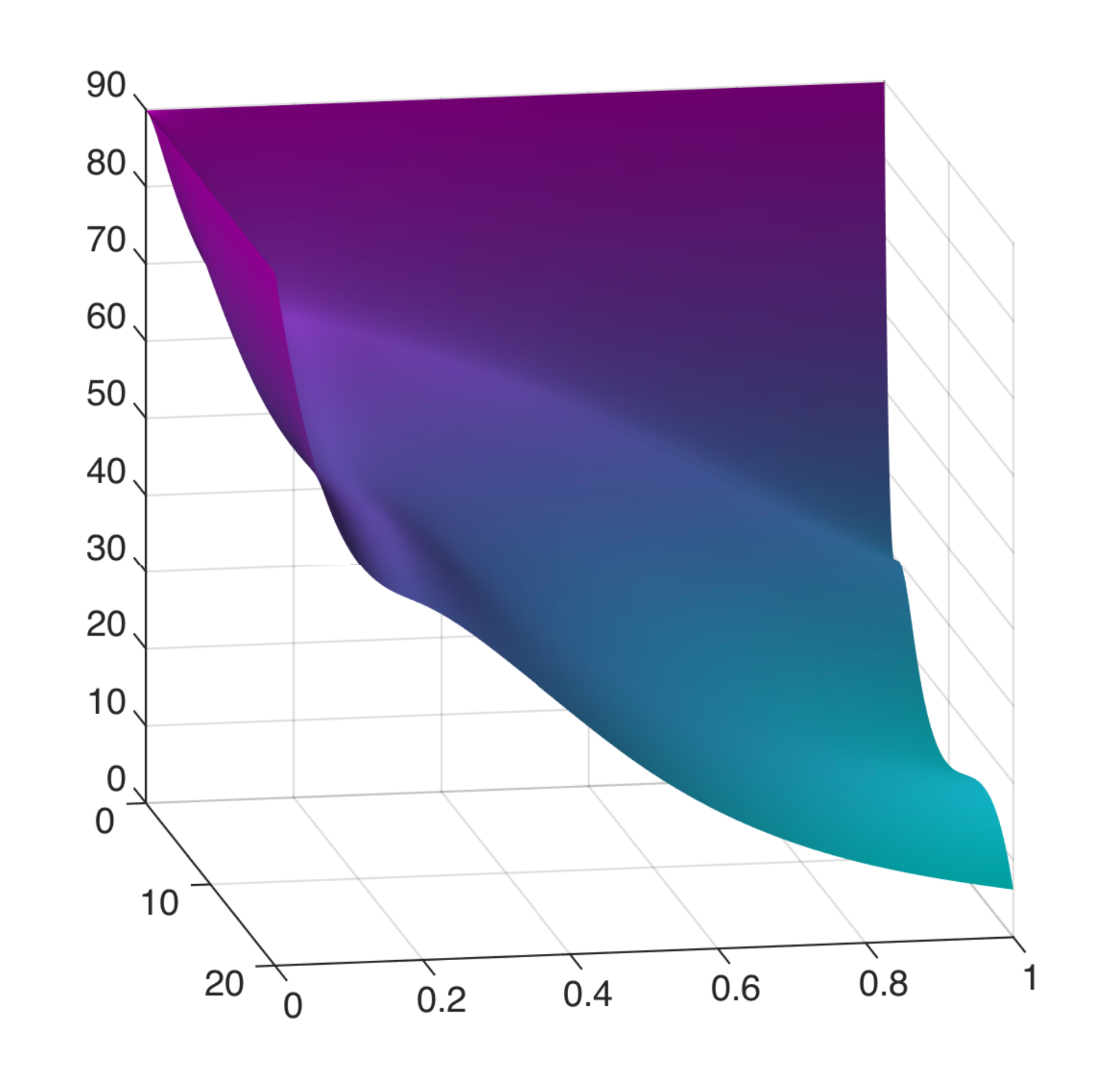}
		\put(26,-1.5){{\rotatebox{1.5}{\sffamily \scalebox{0.7}{Spatial Attack Budget} $\epsw$}} } 
		\put(40,-8){{\rotatebox{1.5}{\sffamily \scalebox{0.7}{(a)}}}} 
		\put(-5,45){\textcolor{black}{{\scalebox{0.7}{\rotatebox{90}{\sffamily  Accuracy (\%)}}}}}
		\put(-7,22){\textcolor{black}{{\scalebox{0.68}{\rotatebox{-52.5}{\sffamily  PGD steps}}}}}
	\end{overpic}
	\hspace{0.1in}
	\begin{overpic}[width=1.4in]{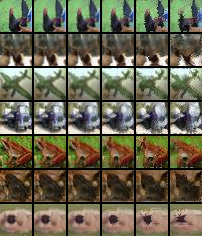}
		\put(0,-2){\vector(1,0){86}}
		\put(70,-8){\scalebox{0.8}{$\epsw\uparrow$}} 
		\put(40,-8){{\rotatebox{1.5}{\sffamily \scalebox{0.7}{(b)}}}} 
	\end{overpic}
	\vspace{0.01in}
	\caption{{Performance of conventional pixel attack-based robust model against spatial attacks.} (a) the  accuracy of a robust model~\cite{madry2017towards} under different spatial attack budgets $\epsw$ using different number of PGD steps. (b) spatial adversarial images with increasing budgets on CIFAR10 using 20  PGD steps for the robust model.}
	\label{fig:adv_images}	
	\vspace{-0.1in}
\end{figure}

In this paper,  we present a joint spatial-pixel adversarial training approach that improves  model robustness.
The contributions of this paper are summarized as follows:
(\emph{i}) we present a spatial attack approach with explicit budgets and a practical \emph{first-order} algorithm for spatial attack generation (as illustrated in Figure~\ref{fig:adv_images}). The proposed setup could serve as one of the first benchmark for evaluating model robustness against spatial attacks; 
(\emph{ii}) we present a framework and concrete algorithms on how to jointly incorporate both pixel and spatial domains for joint adversarial generation;
(\emph{iii})~we develop a joint adversarial training approach that can effectively leverage the joint attacks and improve the models robustness \wrt pixel, spatial and joint attacks; the proposed approach lead to models that achieve state-of-the-art performance in terms of both pixel and spatial attacks.

\section{Pixel Adversarial Training}\label{sec:pixel}

We  review briefly  the standard adversarial training method based on pixel attacks~\cite{FGSM, madry2017towards}, which improves model robustness by solving the following minimax problem
\begin{equation}\label{eq:org_adt_formulation}
\min_{\tt}\{\mathbb{E}_{(\x,y) \sim D}[\max_{\bar{\x} \in \mathcal{S}_{\x}} \L(\bar{\x}, y;\tt)]\},
\end{equation}
where $\x$ and $y$ denote the original image and label sampled from the dataset ${D}$,
$\bar{\x}$  the adversarially perturbed image, $\L(\cdot)$  the loss function, $\tt$ the network parameter, and $\epsx$ the perturbation budget. The feasible region $\mathcal{S}_{\x}$ is defined as~\footnote{Images are scaled to have pixel values in [-1, 1].}
\begin{equation}\nonumber
\mathcal{S}_{\x}=\{\z\,|\, \ \z \in B(\x, \epsx) \cap [-1, 1]^n\},
\label{s_x}
\end{equation}
where $B(\x, \epsx)=\{\z\,|\,\|\z - \x\|_{\infty} \leq \epsx\}$ denotes the \mbox{$\ell_{\infty}$-ball} with center $\x$ and radius $\epsx$.
In the single sample point case, Eqn.(\ref{eq:org_adt_formulation}) reduces to:
\begin{equation}\label{eq:single}
\min_{\tt}\big[\max_{\bar{\x} \in \mathcal{S}_{\x}} \L(\bar{\x}, y;\tt)\big].
\end{equation}
The inner maximization can be approximately solved by either a one-step approach such as the Fast Gradient Sign Method (FGSM)~\cite{FGSM}, or a multi-step method such as the multi-step projected gradient descent (PGD)~\cite{madry2017towards}
\begin{eqnarray}
\x^{t+1} &\!\!\!\!=\!\!\!\!& \P_{\mathcal{S}_{\x}}\big(\x^t + \alpha \cdot \text{sign}\big(\nabla_{\x}\L(\x^t, y; \tt)\big)\big),
\label{PGD}
\end{eqnarray}
where $\P_{\mathcal{S}_{\x}}(\cdot)$ is a projection operator projecting the input into the feasible region $\mathcal{S}_{\x}$.
In the PGD approach, the estimation is initialized as a random point within a neighborhood of the original image $\x$ (\emph{i.e.} $B(\x, \epsx)$), and then goes through several PGD steps with a step size of $\alpha$ as shown in (\ref{PGD}).

\section{Spatial Attacks: Model and Algorithm}\label{sec:spatial}
Most existing methods as reviewed in the previous section are trying to manipulate the pixel value \emph{directly} and \emph{individually}.
While this ensures the most degrees of operational freedom, the resulting model is less robust in face of spatial attacks~\cite{xiao2018spatially,rot_tran} as shown in Figure~\ref{fig:adv_images}, possibly  due to the challenges brought by the excessive degrees of freedom on finding the proper adversarial example for robust training.
It has been shown in \cite{rot_tran} that simple rotation and translation can be a form of naturally occuring adversarial attack. 
While it is argued that simple translational and rotational transformations are ``semantic-preserving'' operations, the transformations are typically large enough to be noticeable~\cite{rot_tran}.  
\cite{xiao2018spatially} generates attacks by deforming a benign image with a flow-based non-rigid transformation and uses a regularization term to encourage smooth deformation of the images. However, there is no direct constraints on the degrees of deformations applied, which makes it less straightforward to be connected with
a quantifiable measure representing the strength of the attack required for quantitative evaluation.

We instead pose an alternative question: \emph{can we achieve effective spatial attack with an explicit budget constraint such that the transformed images are visually indistinguishable from the original ones?}
We give positive answer to this question and present the details in the sequel.

\subsection{Spatial Transformation-based Attack Model}
Given an original image $\x\in \mathbb{R}^n$, a spatially transformed image can be represented as:
\begin{equation}\label{eq:tranformation}
\tilde{\x}=\T(\x, \om),
\end{equation}
where $\T\!:\! \mathbb{R}^{n} \!\!\times\!\! \mathbb{R}^{n\times 2} \!\! \rightarrow \!\!\mathbb{R}^{n}$ denotes a flow-parameterized warping operator with $\om \in \mathbb{R}^{n\times 2}$ denotes the flow filed.
In order to use back-propagation for optimization, a differentiable warping is used~\cite{stn}. While we are focusing on spatial transformation in this work, this notion of transformation is general and our work could potentially be generalized to other forms of transformations al well. We leave the investigation in this direction to future work.

For adversarial  generation, we maximize the classification loss $\L$ similar to the pixel case, but \wrt the flow field:
\begin{equation}\label{eq:spatial_adv}
{\om}^* = \arg \max_{\om \in \mathcal{S}_{\om}} \L(\T(\x, \om), y;\tt).
\end{equation}
Then the spatial transformation-based adversarial attack can be generated as $\bar{\x}=\T(\x, {\om}^*)$.
Note that compared to Eqn.(\ref{PGD}), which edits pixel values individually and directly,  
here we modify images through  transformations thus impacting  pixel values in an implicit way.

\subsection{Explicit Budget for Spatial Attack}
It is important to have a quantitative budget for easy benchmarking as in the pixel case~\cite{madry2017towards}, rather than indirect measurement such as total smoothness~\cite{xiao2018spatially}. Inspired by many seminal works in pixel case~\cite{FGSM, tramer2017ensemble, athalye2018obfuscated, madry2017towards}, where we direct constrain the change of pixel values, we propose to budget the spatial attacks in terms of the displacement amount. 
More specially, given a vector field $\om=[\u, \v] \! \in \! \mathbb{R}^{n\times 2}$, where $\u \!\in \!\mathbb{R}^n$ and $\v \!\in\! \mathbb{R}^n$ denote the horizontal and vertical components, we constrain the flow $\om$ with an explicit spatial budget $\epsw$ as
\begin{equation}\label{eq:spatial_budget}
\om \!\in\! \mathcal{S}_{\om} \equiv B(\mathbf{0}, \epsw) \triangleq \{\om\,|\,\|\om\|_{2,\infty} \leq \epsw\}.
\end{equation}
$\|\cdot\|_{2,\infty}$ denotes the $\mathcal{\ell}_{2, \infty}$-norm and is defined as follows for a general matrix $\M \in \mathbb{R}^{n\times k}$
\begin{eqnarray}\nonumber
\big\Vert \M \big\Vert_{2, \infty} \triangleq \big\Vert \big[\Vert\m_1\Vert_2, \cdots, \Vert\m_i\Vert_2, \cdots\big]^T \big\Vert_{\infty}, \;  1 \le i\le n, 
\end{eqnarray}
where $\m_i$ denotes the $i$-th row of $\M$.
Intuitively speaking, Eqn.(\ref{eq:spatial_budget}) implies that the permissible  flow field cannot have any flow vectors with a length that is larger than $\epsw$.
This notion of explicit budget is crucial for measuring the robustness of different models, thus is one essential prerequisite for designing algorithms on improving model robustness. 
The relationship of varying test time spatial budget and the  accuracy of a robust model~\cite{madry2017towards} on CIFAR10 is depicted in Figure~\ref{fig:adv_images}.
We empirically observed that setting $\epsw$  to one percent the size of the image leads to a reasonable  trade-off between visual quality and attack effectiveness. For CIFAR images of size $32\!\times\! 32$, this implies $\epsw \!\approx\! 0.3$.

\subsection{Generalized Gradient Sign Method}
Towards the goal of robust training,  an efficient algorithm for solving Eqn.(\ref{eq:spatial_adv}) is required  before it can be integrated into  adversarial training.
Some solutions such as grid search~\cite{rot_tran} or L-BFGS method~\cite{xiao2018spatially} are either less scalable or too expensive thus are less appropriate for adversarial training.
More specifically, 
the non-parametric form of $\om$ renders the grid search method used in \cite{rot_tran} impractical. And our goal of incorporating the generation into training pipeline also favors less on  more expensive methods.
Along this line of reasoning, we propose to solve Eqn.(\ref{eq:spatial_adv}) using a  first-order method:
\begin{equation}\label{eq:gradient_descend}
{\om}^{t+1} = \om^{t}  + \alpha\cdot \nabla_{\om}\L(\T(\x, \om), y;\tt).
\end{equation}

To be efficient and effective, both FGSM and PGD method need to incorporate the $\text{sign}(\cdot)$ operator for efficiency in pixel attack. Here for the spatial attack, we would like to have a similar mechanism.
By viewing $\text{sign}(\cdot)$  as a scalar normalization function, \emph{i.e.}, $\text{sign}(x)=\frac{x}{\Vert x\Vert}$, we can define a ``generalized sign'' operator for vectors as:
\begin{equation}
\overrightarrow{\text{sign}}([u, v]) = \frac{[u, v]}{\sqrt{u^2 + v^2}}.
\end{equation}
When $\overrightarrow{\text{sign}}(\cdot)$ takes a flow field $\om$ as input, it operates on each row vector separately.
While the  scalar $\text{sign}(\cdot)$  in the pixel case normalized the scalar value and keeps the sign, the vector $\overrightarrow{\text{sign}}(\cdot)$ normalized the vectors while retaining their original directions.

We can now complete the Generalized Gradient Sign method. 
Firstly, for spaital perturbation, we also need to start  with a random perturbation in the parameter space similar to the pixel case in order to better explore the $\epsw$-ball around the initial all-zero flow field. 
After obtaining the gradient of the flow field as in Eqn.(\ref{eq:gradient_descend}), we apply
the vector  $\overrightarrow{\text{sign}}(\cdot)$ operator to the gradient, and then use it to take an ascend step scaled by a steps size $\alpha$ for updating the flow field.
We then project the flow updated with the gradient to the feasible region $\mathcal{S}_{\om}$ specified by the given budget.
The full procedure  is summarized in Algorithm~\ref{alg:spatial_adv_first_order}.
Note that in the case of multi-step PGD, it is important to accumulate the changes in the transformation parameter $\om$ as shown in Algorithm~\ref{alg:spatial_adv_first_order}, rather than applying transformations to a resulting image accumulatively , which will lead to distorted blurry results due to accumulated transformation errors.

\begin{algorithm}[tb]
	\caption{Generalized Gradient Sign Method}
	\label{alg:spatial_adv_first_order}
	\begin{algorithmic}
		\STATE {\bfseries Input:} image $\x$,  loss $\L$, step $m$, step size $\alpha$, budget $\epsw$
		\STATE Initialize flow $\om^0 \sim B(\0, \epsw)$
		\FOR{$t=1$ {\bfseries to} $m$}
		\STATE $-$ $\bar{\x} = \T(\x, \om^{t-1})$
		\STATE $-$ $\bar{\om}^{t} = \om^{t-1}  + \alpha \cdot \overrightarrow{\text{sign}}(\nabla_{\om}\L(\bar{\x}, y;\tt))$
		\STATE $-$ $\om^{t} = \P_{\mathcal{S}_{\om}}(\bar{\om}^{t})$
		\ENDFOR
		\STATE {\bfseries Output:} adversarial image $\T(\x, \om^{m})$.
	\end{algorithmic}
\end{algorithm}

\subsection{The Effectiveness of First-Order Spatial Attack}
Although the spatial attacks have much less degrees of freedom compared with the pixel attacks, it can attack a  model trained with clean images effectively with a high success rate (\emph{c.f.}~Table~\ref{tab:cifar10}).
Moreover, it can also attack a pixel robustified models with high success rate.
For example, it can reduce the performance of a robust model trained with the state-of-the-art method~\cite{madry2017towards} effectively as shown in Figure~\ref{fig:adv_images}.
It is observed from Figure~\ref{fig:adv_images} that even at a very low budget range (\emph{e.g.} $\epsw \!\le\! 0.5$), the first-order attack can significantly reduce the accuracy rate of a pixel-robustified model~\cite{madry2017towards} while maintaining a
high resemblance of the original image.
Although this degree of resemblance decreases when moving towards the high budget region (\emph{e.g.} $\epsw\!\rightarrow\!1$) as there is no explicit smoothness penalty over the flow field, the perturbed images still preserve the global structure thus the major information~\cite{orderless_image}. 
Furthermore, we observe that under a fixed budget, the proposed method can reduce the accuracy of the robustified model significantly after  a single step and the 
the attacking strength  increases with increased number of PDG steps.

In summary, we have empirically observed the effectiveness of the proposed first-order approach for generating spatial attacks. Different from the observations in \cite{rot_tran}, we  found that the  first-order optimization method is fairly effective in generating adversarial spatial attacks. This enables us to further utilize it for adversarial model training as detailed in the next section.

\section{Joint Adversarial Training}
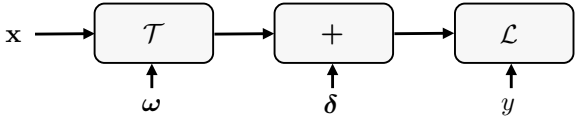
\begin{figure}
 	\centering
 	\begin{overpic}[width=3in]{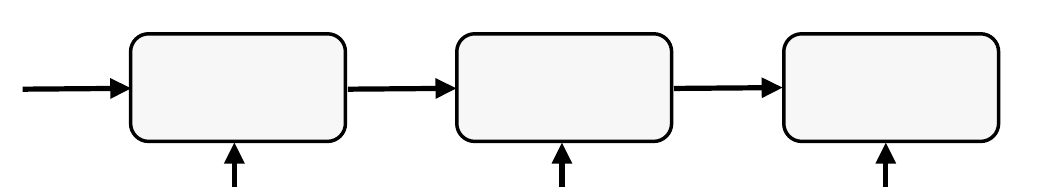}
 		\put(-3,8.5){$\x$}
 		\put(21,8.5){$\T$}
 		\put(52,8.8){$\boldsymbol{+}$}
 		\put(20.8,-3.5){$\om$} 
 		\put(52.5,-3.8){$\dt$}    
 		\put(83.3,8.3){$\L$}  
 		\put(83.5,-3.3){$y$} 
 	\end{overpic}
 	\vspace{0.15in}
 	\caption{Computational graph for joint adversarial attack and training. $\T$ denotes the spatial transformation operator while $+$ corresponds to pixel-wise additive attacks. For generating joint adversarial attacks, instead of optimizing over $\x$ directly, we optimize over $\om$, $\dt$ and then use them to generate the perturbed image.}
 	\label{fig:comp_graph}
 \end{figure}

In order to jointly incorporate both spatial and pixel perturbations, we first present a re-parameterization of the adversarial image in the pixel attack case as follows:
\begin{equation}\label{eq:reparam}
\bar{\x}=\x + \dt.
\end{equation}
Based on this re-parameterization, we can conventionally  switch from optimizing over $\bar{\x}$ as in Eqn.(\ref{eq:org_adt_formulation}) to optimizing over $\dt$. While this is insignificant when considering pixel attacks only, it facilities our derivation in the presence of multiple attacks. 
Concretely, when incorporating  both spatial and pixel transformations, we have the following model for joint spatial-pixel adversarial attack generation: 
\begin{equation}\label{eq:reparam_trans}
\bar{\x}=\T(\x, \om) + \dt.
\end{equation}
The corresponding computational graph is shown in Figure~\ref{fig:comp_graph}.
Based on this, we can generate the adversarial image by optimizing over both $\om$ and $\dt$ through a proper loss function $\L$. 
Given Eqn.(\ref{eq:reparam_trans}), our formulation for the joint adversarial training task is as follows:
\begin{equation}\label{eq:JAT}
\min_{\tt}\big[\max_{\om \in \mathcal{S}_{\om}, \dt \in \mathcal{S}_{\dt}} \L\big(\T(\x, \om) + \dt, y;\tt\big)\big].
\end{equation}
The feasible region $\mathcal{S}_{\dt}$ is defined as
\begin{equation}\nonumber
\mathcal{S}_{\dt}=\{\dt\,|\, \ \dt + \T(\x, \om) \in B^{\mathcal{T}}(\x, \epsx) \cap [-1, 1]^n\},
\label{s_x}
\end{equation}
where $B^{\mathcal{T}}(\x,\epsx)=\{\z\,|\, \|\z - \mathcal{T}(\x, \om)\|_{\infty} \leq \epsx\}$, which essentially means the element $\z$ is in the ball when it is close to the transformed clean image $\mathcal{T}(\x, \om)$ in terms of $\mathcal{\ell}_{\infty}$-norm.
The adversarial example is generated by solving the inner problem of Eqn.(\ref{eq:JAT}) optimizing over both $\om$ and $\dt$.
We will present a practical algorithm for this in the sequel.

\subsection{Joint Attack via Double-Pass Algorithm}\label{subsec:joint_attack}
To solve for the inner maximization problem of Eqn.(\ref{eq:JAT}), we propose a double-pass algorithm for properly handling the interactions between the two types of attacks. 
The detailed algorithm for joint spatial-pixel adversarial attack generation is summarized in Algorithm~\ref{alg:joint_adv}.
We start from random points for both $\om$ an $\dt$ in their respective $\epsilon$-balls. 
In the first pass, we forward the clean image $\x$ through the adversarial generation model Eqn.(\ref{eq:reparam_trans}) with the current estimations of $\om$ and $\dt$, obtaining the adversarial image $\bar{\x}$.
Then the gradient of the loss \wrt $\om$ is computed based on $\bar{\x}$ and the generalized gradient sign approach (Algorithm~\ref{alg:spatial_adv_first_order}) is applied for updating $\om$.
In the second pass, we forward the clean image again through the adversarial generation process with the updated $\om$ and perform a gradient update for $\dt$ based on the gradient of the current loss \wrt it.
This process can be repeated for multiple steps and each step  can  be essentially viewed as a variant of the alternating minimization approach at instance level.
The final adversarial image is generated as $\T(\x, \om^{m}) + \dt^{m}$, where $m$ denotes number of steps.

\begin{algorithm}[tb]
	\caption{Double-Pass Joint Adversarial Attack}
	\label{alg:joint_adv}
	\begin{algorithmic}
		\STATE {\bfseries Input:} image $\x$, label $y$, loss $\L$, step $m$, step size $\alpha$, $\beta$, 
		\STATE \quad\qquad budget $\epsw$, $\epsx$
		\STATE Initialize $\om^0 \sim B(\0, \epsw)$, $\dt^0 \sim B(\0, \epsx)$
		\FOR{$t=1$ {\bfseries to} $m$}
		\STATE $-$ $\bar{\x} = \T(\x, \om^{t-1}) + \dt^{t-1}$  // first-pass
		\STATE $-$ $\bar{\om} = \om^{t-1}  + \alpha \cdot \overrightarrow{\text{sign}}(\nabla_{\om}\L(\bar{\x}, y;\tt))$
		\STATE $-$ $\om^{t} = \P_{\mathcal{S}_{\om}}(\bar{\om})$
		\STATE $-$ $\bar{\x}' = \T(\x, \om^{t}) + \dt^{t-1}$   // second-pass
		\STATE $-$ $\bar{\dt} = \dt^{t-1}  + \beta \cdot \text{sign}(\nabla_{\dt}\L(\bar{\x}', y;\tt))$
		\STATE $-$ $\dt^{t} = \P_{\mathcal{S}_{\dt}}(\bar{\dt})$
		\ENDFOR
		\STATE {\bfseries Output:} joint adversarial image $\T(\x, \om^{m}) + \dt^{m}$.
	\end{algorithmic}
\end{algorithm}

\subsection{Practical Joint Adversarial Training}
Training robust models that is 
resistant to both spatial and pixel attacks boils down to solving a minimax problem as in  Eqn.(\ref{eq:JAT}).
We approximately solve Eqn.(\ref{eq:JAT}) by replacing the original clean training images with the joint adversarially perturbed images obtained through the inner problem, and then performing a conventional training of the model using the perturbed images as done in conventional adversarial training~\cite{FGSM,madry2017towards}. 

The inner optimization problem corresponds to the attack generation problem and  we have presented a first-order approach for joint attack generation in Section~\ref{subsec:joint_attack} as summarized in Algorithm~\ref{alg:joint_adv}.
In practice, we use a targeted approach for generating the adversarial images.
We replace the ground-truth $y$ with another label randomly sampled from the set of labels excluding the correct one and flip the sign of the gradient.
A variant of label smoothing is also used~\cite{bilateral} for training. The complete procedure for joint adversarial training is summarized in Algorithm~\ref{alg:JAT}.

{\flushleft\textbf{Comments.}} We have empirically observed that the joint attack is a type of attack that is stronger than either pure pixel or spatial attacks, as shown in Table~\ref{tab:cifar10}$\sim$\ref{tab:svhn}. 
It can be observed that the pixel robustified models are still vulnerable to spatial attacks. This  suggest that these two types attacks are not entirely in the same space, thus offering complementary strength to some extent.
This is in line with the observation of \cite{rot_tran}.
On the other hand, jointly considering both forms of attacks can further improve the robustness of the model compared with that of the model trained with only one type of attacks. This implies that the two forms of attacks share part of a common space where they can interact, which is different from the observation in \cite{rot_tran} that they are orthogonal to each other.
Our observation is reasonable and is aligned with the fact that both are eventually impacting the pixel values, thus indeed sharing some common operational ground.
Finally, it is worthwhile to explain a few points about the proposed double-pass algorithm. Firstly, it is different from simply performing pixel adversarial after spatial perturbation as two isolated pieces. Secondly,
the straightforward one-pass approach does not perform well potentially due to the reason that it cannot handle the conflict of two types of attacks properly.
More discussions on these points, including the order of the two types of attacks, will be elaborated in the experiment section.

\begin{algorithm}[tb]
	\caption{Joint Adversarial Training}
	\label{alg:JAT}
	\begin{algorithmic}
		\STATE {\bfseries Input:} training epochs $K$, batch size $N$, learning rate $\rho$,  dataset ${D}$, budget $\epsw$, $\epsx$,
		\FOR{$k=1$ {\bfseries to} $K$}	
			\FOR{$\{\x_j, y_j\}_{j=1}^N$ from the first {\bfseries to} last  batch from ${D}$}	
		\STATE $-$ generate target  $\hat{y}_j \!\!=\!\!\text{rand\_unif}(\{0, 1, \tiny{\cdots}\} \backslash \{y_j\})$
		\STATE $-$ generate jointly perturbed  image batch $\{\bar{\x}_j\}_{j=1}^N$ 
		\STATE \quad according to Algo.~\ref{alg:joint_adv} using targets  $\{\hat{y}_j\}$ 
		\STATE $-$ get $\{\bar{y}_j\}$ with $\bar{y}_j=\text{label\_smooth}(y_j)$
		\STATE $-$  $\tt = \tt  - \rho \cdot (\frac{1}{N}\sum_{j=1}^N\nabla_{\tt}\L(\bar{\x}_j, \bar{y}_j;\tt))$
			\ENDFOR
		\ENDFOR
		\STATE {\bfseries Output:} model parameter $\tt$.
	\end{algorithmic}
\end{algorithm}

\section{Discussions}
We discuss the relationship of the proposed approach with some previous related works  to help understanding the connections with the literature.
{\flushleft \textbf{Pixel Adversarial Attacks and Defences}.} 
Model robustness and adversarial examples have  been investigated under different contexts~\cite{Dalvi,svm}.  Adversarial examples for CNN is investigated in \cite{szegedy2013intriguing}. \cite{goodfellow2014explaining} proposed a efficient   adversarial attack generation approach called fast gradient sign method (FGSM). Many efforts have been devoted in developing variants of attacks~\cite{moosavi2015deepfool,carlini2016towards}.  
Defense against adversarial attacks has also emerged as an important topic and attracted 
many attention~\cite{metzen2017detecting,meng2017magnet, xie2017mitigating,guo2017countering,liao2017defense,samangouei2018defensegan,song2018pixeldefend,prakash2018deflecting,liu2017towards,bilateral}.  Adversarial training~\cite{goodfellow2014explaining, kurakin2016scale, tramer2017ensemble, madry2017towards, athalye2018obfuscated} is one effective defence method against adversarial attacks, 
It formulates the task of robust training as a minimax problem, where the inner maximization essentially generates attacks while the outer minimization corresponds to minimizing the ``adversarial loss'' induced by the inner attacks~\cite{madry2017towards}. It typically utilizes multiple steps for solving the inner problem in order to achieve good performance, which makes the training process time-consuming. 
\cite{bilateral} recently introduces a label adversarial procedure for  adversarial training which achieves state-of-the-art performances with a single-step method, with the incorporation of label perturbation based on a variant of label smoothing, in addition to the conventional image perturbation.
{\flushleft \textbf{Spatial Transformation and Adversarial Attacks}.} 
Spatial transformation has been playing a crucial role in training deep network models. It has been commonly used for augmenting the training data for training deep networks~\cite{Alex}. 
The spatial transformer network has been used to further improve the invariance of the model \wrt spatial transformations of input images~\cite{stn}.
A few recent works investigated the role of spatial transformation in attacks. \cite{rot_tran} showed that simple rotation and translation can perform attack effectively. \cite{xiao2018spatially} used a  flow field to deform a benign image smoothly to generate attacks. We also use a flow field for spatial transformation. But instead of \emph{implicitly penalize} for the flow filed as done in \cite{xiao2018spatially}, we propose to \emph{explicitly constrain} the flow with a budget. Furthermore, we develop a practical first-order approach for efficient adversarial generation, which is more practical compared to grid search~\cite{rot_tran} or L-BFGS~\cite{xiao2018spatially}.
{\flushleft \textbf{Differentiable Renderer}.} 
The proposed joint adversarial generation model resembles a standard image formation process and can actually be regarded as an instance of differentiable renderers~\cite{OpenDR,halide}, which has been used for tackling many tasks including image restoration~\cite{reblur}, 3D shape estimation~\cite{OpenDR} and  model-based reinforcement learning~\cite{render_RL}. Recently a 3D model-based differentiable renderer has been used for generating adversarial images~\cite{render_adv}. 
Here we use the generation model for joint spatial-pixel adversarial generation without the aid of 3D models. Still, this interesting connection sheds lights on possible future directions.


\begin{figure}[t!]
	\centering
	\vspace{0.05in}
	\begin{overpic}[viewport = 18 10 370 290, clip, height=1.25in]{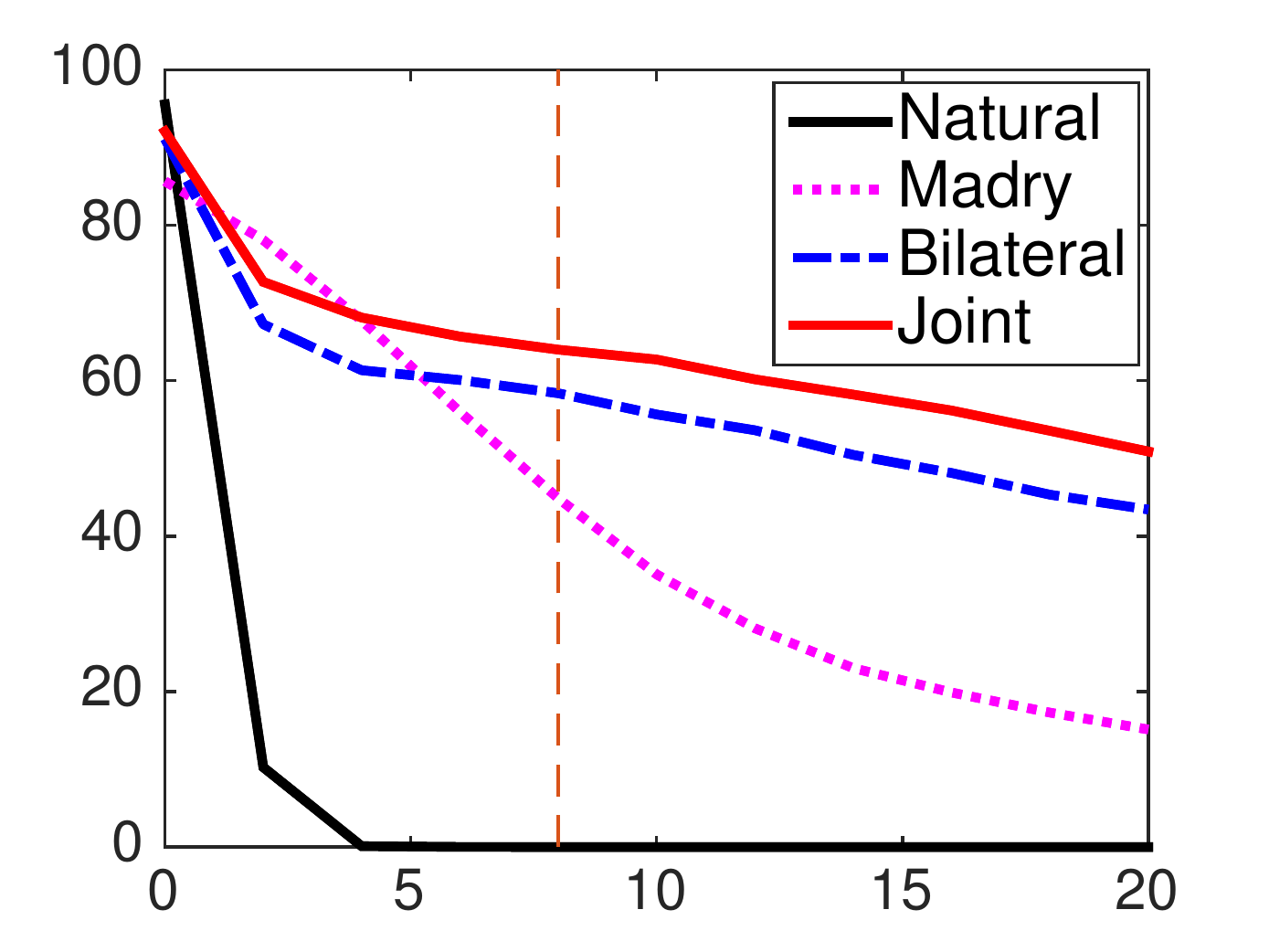}
		\put(-4.5,33){\sffamily \small {\textcolor{black}{{\scalebox{0.8}{\rotatebox{90}{Acc.(\%)}}}}}}
		\put(9,80){\sffamily\small{\textcolor{black}{{\scalebox{0.8}{Performance under Pixel Attack}}}}}
		\put(36,-7){\sffamily\small{\textcolor{black}{{\scalebox{0.8}{Eval. Budget }$\epsx$}}}}
	\end{overpic}
	\hspace{0.04in}
	\begin{overpic}[viewport = 18 10 370 290, clip, height=1.25in]{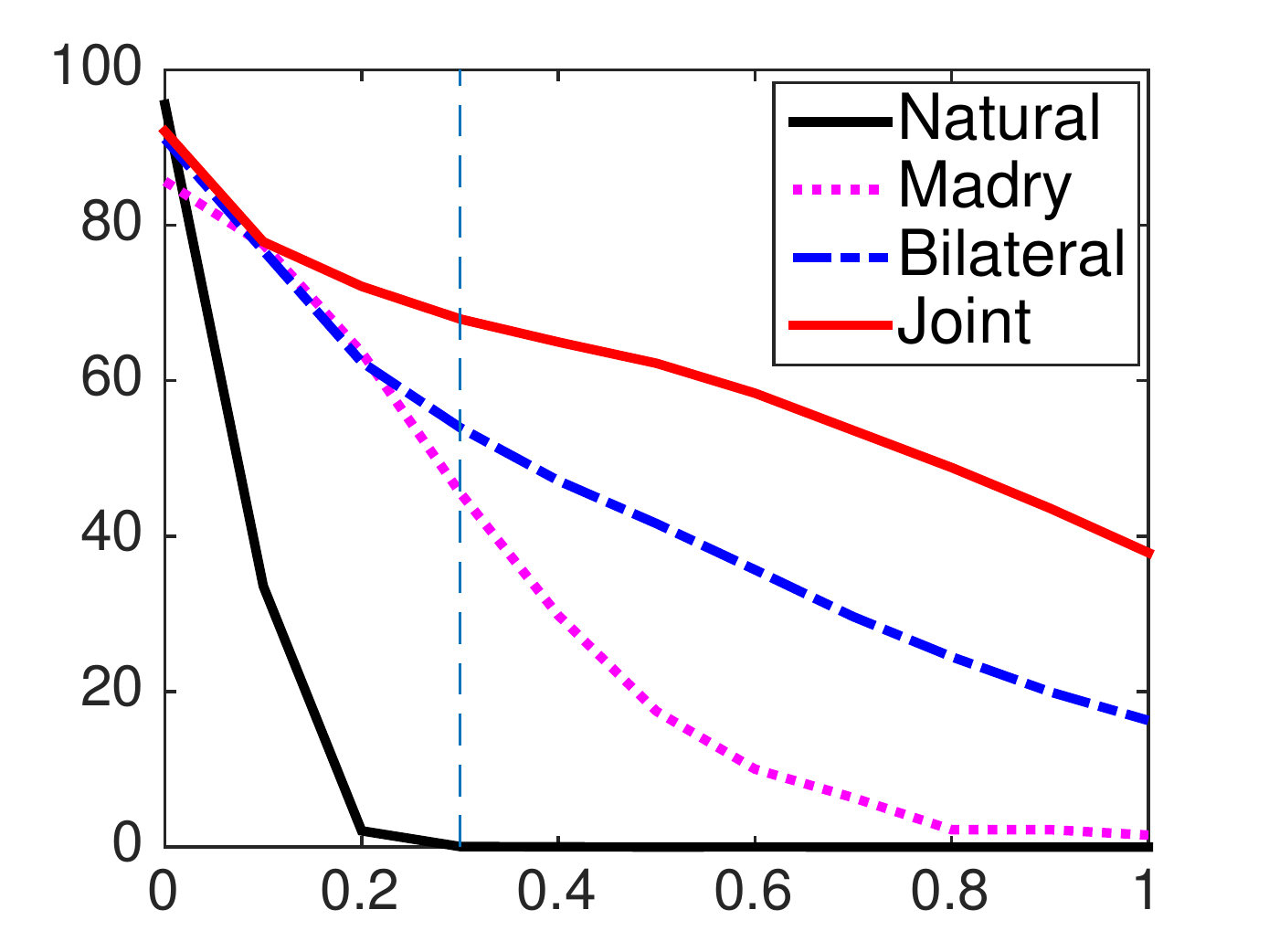}
		\put(-4.5,33){\sffamily \small {\textcolor{black}{{\scalebox{0.8}{\rotatebox{90}{Acc.(\%)}}}}}}
		\put(36,-7){\sffamily\small{\textcolor{black}{{\scalebox{0.8}{Eval. Budget }$\epsw$}}}}
		\put(5,80){\sffamily\small{\textcolor{black}{{\scalebox{0.8}{Performance under Spatial Attack}}}}}
	\end{overpic}
	\vspace{0.1in}
	\caption{Robustness of different models at different pixel attack levels  (20 steps PGD) on CIFAR10. The vertical dashed lines represent the budgets used during training. It is observed that the proposed approach can improve its robustness \wrt both pixel and spatial attacks compared to the state-of-the-art methods.}
	\label{fig:attack_budgets_plot}	
\end{figure}

\begin{figure*}[t!]	
	\centering
	\includegraphics[height=1.18in]{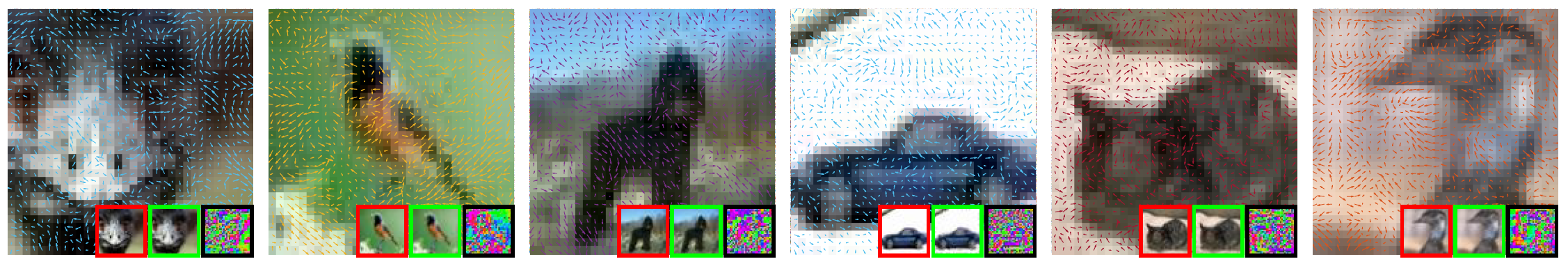}
	\caption{Visualization of the flow field generated by the proposed approach. The images enclosed with red, green and black lines correspond to the original image, perturbed image with proposed spatial attack and the corresponding flow field image.}
	\label{fig:flow_viz}
	\vspace{-0.15in}	
\end{figure*}

\section{Experiments}
Our implementation is based on PyTorch\footnote{\url{https://github.com/pytorch}} and the code to reproduce our results will be available on the project page.\footnote{\url{https://github.com/author/joint_adverse}}
Extensive experiments are conducted on various datasets including \mbox{CIFAR10}~\cite{krizhevsky2009learning},  \mbox{CIFAR100}~\cite{krizhevsky2009learning} and SVHN~\cite{netzer2011reading}.   We perform comparisons with several state-of-the-art adversarial training methods, as well as a number of variants of our proposed approach as detailed in the sequel.
{\flushleft \textbf{Baselines}.}  We categorize all the methods to several categories according to the type of attack used during training. For each category there are \salg{one-step} and \salg{multi-step} variants apart from the standard training method:
\begin{itemize}[leftmargin=10pt]
	\itemsep-0.3em
	\item \alg{Natural}: standard  training using the original images.
	\item \alg{Pixel}: pixel-based robust training methods.  \salg{multi-step} variant corresponds to \salg{Madry} method~\cite{madry2017towards} taking multiple PGD steps with random start. \salg{one-step} version corresponds to \salg{Bilateral} method~\cite{bilateral} using  gradient descend with targeted adversarial image generation and a variant of label smoothing. 
	\item \alg{Spatial}: the approach that uses the proposed spatial attacks (Algorithm~\ref{alg:spatial_adv_first_order}) for model training.
	\item \alg{Joint}: the proposed approach that leverages joint spatial and pixel attacks robust training (Algorithm~\ref{alg:JAT}).
\end{itemize}
{\flushleft \textbf{Test Time Attack}.} For  robustness evaluation, we use pixel budget $\epsx\!\!=\!\!8$ and spatial budget $\epsw\!\!=\!\!0.3$.
We used FGSM~\cite{FGSM}, and multi-step PDG$S$~\cite{madry2017towards} as two representative pixel-based attack methods where $S$ denotes the number of steps used.
We also use our developed Algorithm~\ref{alg:spatial_adv_first_order} as a spatial attack method (denoted as ``Spatial''), and the joint spatial-pixel attack method presented in Algorithm~\ref{alg:joint_adv} as a joint attack method (denoted as ``Joint'').
FGSM used a step size of $8$.
All other attacks take $20$ steps, with pixel step size of $2$ and spatial step size of $0.15$.
To generate strong attacks for evaluating robustness, we always use non-targeted attack, and use random start for all methods except for FGSM. 
\begin{table}[t]
	\centering
	\tabcolsep=0.1cm
	\begin{tabular}{ |c|c||c|c|c|c|c|}
		\hline
		\multicolumn{2}{|c||}{\multirow{2}{*}{Models Acc.(\%)}}
		& \multirow{2}{*}{Pristine}
		& \multicolumn{4}{c|}{Test Time  Attack} \\
		\cline{4-7}
		\multicolumn{2}{|c||}{}  & &  \small{FGSM} & \small{PGD20} & \small{Spatial}  & \small{Joint}\\
		\hline \hline
		\multicolumn{2}{|c||}{\salg{Natural}} & {95.6} & 36.9& 0.0 & 0.1 & 0.0  \\
		\hline \hline
		\multirow{3}{*}{\parbox{1cm}{\centering \salg{multi step}}} & \salg{Pixel} & 85.7 & 54.9 & 44.9 & 45.6 & 11.6 \\
		\cline{2-7}
		&  \salg{Spatial} & 91.0 &  57.0  & 18.7  & {70.8}  &  13.4  \\
		\cline{2-7}
		& \salg{Joint}  &  76.4 &	55.8&	50.4&	60.2&	29.6  \\ 
		\hline    \hline      
		\multirow{3}{*}{\parbox{1cm}{\centering \salg{one step}}}	& \salg{Pixel} & 91.1& {90.5} & 58.4 & 53.7 & 35.8 \\ 
		\cline{2-7}
		& \salg{Spatial} & 94.1 & 73.2 & 35.8 & 65.7 & 30.5 \\
		\cline{2-7}
		& \salg{Joint}  & 92.1 &{88.7}  &{64.0}  & {68.0}  & {53.1} \\
		\hline
	\end{tabular}
	\caption{Accuracy of different models under different attacks on CIFAR10. \salg{multi-step-Pixel} corresponding to the \salg{Madry} methopd~\cite{madry2017towards} and the \salg{one-step-Pixel} corresponding to the \salg{Bilateral} method~\cite{bilateral}.}
	\label{tab:cifar10}
\end{table}	
{\flushleft\textbf{Implementation Details.}} 
We use Wide ResNet (WRN-28-10)~\cite{zagoruyko2016wide}  following~\cite{madry2017towards}. 
For $\epsw$, we set it to be $1\%$ of image size, which corresponds to  $\epsw\!\!=\!\!0.3$ pixel for CIFAR10, CIFAR100 and SVHN datasets.
The initial learning rate   is $0.1$ for CIFAR and $0.01$ for SVHN.
We set the number of epochs for all the multi-step methods as $100$ with transition epochs as $\{60, 90\}$ as we empirically observed the performance of the trained model stabilized before $100$ epochs. 
When incorporating spatial attacks, $300$ epochs with transition epochs $\{100, 150\}$ is used as we empirically observed it helps with the model performance, possibly due to the increased variations of data.
The learning rate decay factor is $0.1$.
\alg{Madry} method~\cite{madry2017towards} takes multi-step PGD  with the number of steps as $7$ and step size as $2$ to be compatible with the literature~\cite{madry2017towards}. 
For other multi-step methods, we use the number of steps as $5$, $0.1$ for the spatial step size $\alpha$ and $2$ for the pixel step size $\beta$ for training.
The transformation operator is implemented as a differentiable bilinear warping following~\cite{stn}.

\subsection{CIFAR10}
We conduct experiments on CIFAR10~\cite{krizhevsky2009learning} and compare the performance of the proposed method with a number of baseline methods in this section.
CIFAR10~\cite{krizhevsky2009learning} is a popular dataset that is widely use in adversarial training literature~\cite{madry2017towards,rot_tran,bilateral} with $10$ classes, $5$K training images per class and $10$K test images.
We perform standard data augmentation  including random crops with $4$ pixels of padding and random horizontal flips~\cite{krizhevsky2009learning}.
The results are summarized in Table~\ref{tab:cifar10}.

We have several interesting observations: 
\emph{(i)}~pixel-based robust models (\salg{multi-step-Pixel} \emph{i.e.} \salg{Madry}~\cite{madry2017towards}, \salg{one-step-Pixel} \emph{i.e.} \salg{Bilateral}~\cite{bilateral}) have certain level of robustness to the spatial attacks, but are still less robust to spatial attacks compared with the spatial robust models.
On the other hand, spatial robust models improve the robustness \wrt to spatial attacks significantly, and also improve the robustness \wrt pixel attacks at the same time, although it is still lacked compared with pixel robust models. This  suggests that the pixel and spatial attacks neither work in two totally independent spaces nor in a fully aligned space; instead, they share a common subspace at least in part so that they can contribute to improving  robustness \wrt each other. 
This contributes a complementary view to the one in~\cite{rot_tran} that the two types of attacks (with a rigid spatial transformation) seems orthogonal to each other.
\emph{(ii)}~the proposed \salg{Joint} approach can further improve the model robustness compared with {state-of-the-art} methods, either under the \salg{multi-step} or \salg{single-step} settings.
For example, when trained with one-step joint attacks (\salg{one-step-Joint}), the model has a level of robustness on-par or even better than the models specifically trained to be resistant to a particular type of attacks (\emph{e.g.}, \salg{one-step-Pixel}, \salg{one-step-Spatial}).
The improvement is more pronounced when considering evaluation with the joint spatial-pixel attacks, indicating the effectiveness of the proposed method.
We will use the one-step variant for the proposed method in the following experiments.

We further compare different models under different attack levels (budgets) and the results are shown in Figure~\ref{fig:attack_budgets_plot}. It can be observed that the \salg{Joint} model trained with the proposed approach outperforms other  methods under both pixel and spatial attacks. Moreover, it is also interesting to note that although the model  is trained under a fixed budget (marked with a dashed vertical line), it generalizes to a range of  attacks with varying budgets reasonably well.
\begin{table}[t!]
	\centering
	\tabcolsep=0.1cm
	\begin{tabular}{ |c||c|c|c|c|c|}
		\hline
		{\multirow{2}{*}{Models Acc.(\%)}}
		& \multirow{2}{*}{Pristine}
		& \multicolumn{4}{c|}{Test Time Attack} \\
		\cline{3-6}
		&&  \small{FGSM} & \small{PGD20} & \small{Spatial}  & \small{Joint}\\
		\hline      \hline
		\salg{Cascade} &{92.3}&	92.2&	48.7&	49.4&	31.8  \\ 
		\cline{1-6}
		\salg{One-pass} &{91.8} & {95.5} &	{32.6}  &	{44.9} & {23.8} \\
		\cline{1-6}
		\salg{Joint}  & 92.1 &88.7  &{64.0}  & {68.0}  & {53.1} \\
		\hline
	\end{tabular}
	\caption{Comparison of different solutions integrating both spatial and pixel attacks on CIFAR10 ($\epsx\!=\!8, \epsw\!=\!0.3$).} 
	\label{tab:one_double_pass}
\end{table}
\subsection{Comparison of Integration Approaches: \\Cascade, One-Pass and Double-Pass}
We investigate several variants of ways to integrate and solve spatial and pixel attacks for robust model training. \emph{(i)}~\salg{Cascade} is a baseline approach that simply cascades spatial and pixel attacks as two isolated modules. 
\emph{(ii)}~\salg{One-pass} is a baseline approach that uses a single forward-backward pass for updating both the spatial transformation and pixel additive values. 

It can be seen from the results in Table~\ref{tab:one_double_pass} that neither \salg{Cascade} nor \salg{One-pass} cannot fully leverage the spatial and pixel attacks to improve the model robustness.
In fact, it is interesting to observe that the natural idea of naively cascading two types of attacks cannot show clear gain and sometimes compromises the robustness.
Also, the straightforward one-pass  approach  does not work well, implying the importance of handling the two types of attacks properly. 
\salg{Joint} method  with the double-pass algorithm can effectively improve the model robustness and outperforms \salg{Cascade} and \salg{One-pass} methods by a large margin.

 \begin{table}[t!]
 	\centering
 	\tabcolsep=0.05cm
 	\begin{tabular}{ |c||c|c|c|c|c|}
 		\hline
 		\multirow{3}{*}{Model Acc.(\%)}
 		& \multirow{3}{*}{Pristine}
 		& \multicolumn{4}{c|}{Test Time Attack} \\
 		\cline{3-6}
 		&   & \multirow{2}{*}{\small Pixel} & \multirow{2}{*}{\small Spatial}  & \multicolumn{2}{c|}{\small Joint}\\
 		\cline{5-6}
 		&   &                        & & {\scriptsize{Spatial-Pixel}}& {\scriptsize Pixel-Spatial}\\
 		\hline \hline
 		{\salg{Joint-SP}} & 92.1  &{64.0}  & {68.0}  & 53.1 & 48.5 \\
 		\hline 
 		\salg{Joint-PS} & 90.4& 64.9 &	67.6&	59.8	& 62.1   \\
 		\hline
 	\end{tabular}
 	\caption{The impact of spatial-pixel attack ordering of the proposed method on model robustness  on CIFAR10 ($\epsx\!=\!8, \epsw\!=\!0.3$).}
 	\label{tab:ordering}
 \end{table}

 \subsection{The Ordering of Spatial and Pixel Attacks}
 While we have presented the joint adversarial training approach in a spatial-pixel attack ordering as in Eqn.(\ref{eq:reparam_trans}).
 The derivations are general and it is straightforward to be generalized to the pixel-spatial ordering. Here we experiment with two variants of our proposed approach and investigate the  impacts of ordering on the model robustness.
 We use \alg{Joint-SP} and \alg{Joint-PS} to denotes the models trained following  the proposed approach with spatial-pixel and pixel-spatial ordering respectively.
 The results are reported in Table~\ref{tab:ordering}.
 It is observed that models trained under different attack orders might have slightly different performance under a particular attack but overall deliver comparable robustness performances.  More details on this including the derivations will be provided in the supplementary file.
 
 \begin{table*}[t!]
 	\centering
 	\tabcolsep=0.235cm
 	\begin{tabular}{ |p{1.5cm}|c|c|c|c|c|c|c|c|c|c|c|c|c|}
 		\hline
 		\multirow{2}{*}{\parbox{1.5cm}{\centering Train Budgets}} & \small{Pixel} $\epsx$ & \multicolumn{3}{c|}{2}  & \multicolumn{3}{c|}{4}  & \multicolumn{3}{c|}{6}  & \multicolumn{3}{c|}{8} \\
 		\cline{2-14}
 		&	\small{Spatial} $\epsw$  & 0.1 & 0.2 &  0.3  & 0.1 & 0.2 &  0.3 & 0.1 & 0.2 &  0.3 & 0.1 & 0.2 &  0.3 \\
 		\hline \hline
 		\multirow{5}{*}{\parbox{1.5cm}{\centering Test Time Attack}} & \small{Pristine} &92.2 &	90.5&	90.2&	91.3&	89.1&	88.0&	90.1&	88.7&	87.3&	{94.2}&	92.5&	92.1   \\
 		\cline{2-14}
 		&\small{PDG20}  & {68.7}&	65.9&	61.6&	65.4&	61.4&	57.8&	64.5&	59.2&	57.5&	65.2&	62.3&	64.0 \\
 		\cline{2-14}
 		&\small{PDG100}  &{67.5}&	65.3&	60.3&	64.7&	60.7&	57.0&	64.0&	58.5&	57.1&	54.4&	50.2&	52.7  \\
 		\cline{2-14}
 		&\small{Spatial}  &68.0&	67.5&	67.2&	64.0&	64.0&	63.2&	62.7&	61.4&	61.7&	68.0&	{69.1}&	68.0  \\
 		\cline{2-14}
 		&\small{Joint}  &61.53&	{62.8}&	57.3&	59.4&	57.1&	53.6&	58.8&	54.0&	53.5&	46.4&	46.7&	53.1  \\
 		\hline
 	\end{tabular}
 	\caption{Evaluation accuracy of models trained with the proposed \salg{Joint} method under different training budgets $(\epsx, \epsw)$ on CIFAR10.}
 	\label{tab:train_budget}
 \end{table*}

\subsection{The Impact of Training Budgets}
We investigate the impacts of training budgets in this section and the results are summarized in Table~\ref{tab:train_budget}.
It is observed that the performance of the proposed method is relatively stable \wrt the variations of budgets, and it seems to be beneficial to use  smaller budgets for training.
For example, our model achieves the best result (68.7\% under PGD20 on CIFAR10) with $\epsx\!\!=\!2$ and $\epsw\!\!=\!\!0.1$. 
The purpose of this set of experiments is to show that different training budgets do have different impacts on the final model robustness. 
However we do not aim to select the best performing model by grid-searching over the combinations of training budgets and we will use a training budgets the same as the test budgets by default in the following experiments， while
bearing in mind that the numbers for model robustness could be further increased by tuning  them.
What is the most appropriate training budgets and how to determine it is an open problem and worth to be investigated in the future work.

\begin{table}[h]
	\centering
	\tabcolsep=0.1cm
	\begin{tabular}{ |c|c|c|c|c|c|}
		\hline
		\multicolumn{2}{|c|}{Black-box Attack}	&  \small{FGSM}	&  \small{PDG20}  & \small{Spatial}  & \small{Joint} \\
		\hline \hline
		\multirow{2}{*}{\parbox{2cm}{\centering  Attack Gen. Model}}	&Undefended  &88.7&	90.7&	90.4&	90.5  \\
		\cline{2-6}
		& Siamese & 88.7& 82.4	 &85.2	&79.8 \\
		\hline
	\end{tabular}
	\caption{Accuracy of the \salg{Joint} approach under \emph{black-box} attack on CIFAR10 dataset.  Model used for black-box attacks (Attack Gen. Model): ``Undefended''-the model trained with clean images, 
		``Siamese''	 another model trained with the \salg{Joint} approach.}
	\label{tab:blackbox}
\end{table}

\subsection{Black-box Attack}
To further evaluate the robustness of the proposed \alg{Joint} approach \wrt \emph{black-box} attacks, we conduct experiments using undefended model and a jointly trained model in another training session for generating test time attacks.
The results are presented in Table~\ref{tab:blackbox}. As demonstrated by the results, the model trained with the  \alg{Joint} approach is robust against various of black-box attacks, verifying that a non-degenerate solution is learned~\cite{tramer2017ensemble}.

\subsection{CIFAR100}
We evaluate the models against white-box attacks on \mbox{CIFAR100} dataset and  with $100$ classes, $50$K training and $10$K test images~\cite{krizhevsky2009learning}.
As shown by the  results in Table~\ref{tab:cifar100}, the proposed approach can effectively  increase the robustness of the model \wrt spatial attack and joint attack as observed for CIFAR10.
Also, it helps to boost the model robustness \wrt pixel attacks as well, which is well-aligned with the analysis in previous sections, that both forms of attacks have some complementary strength that can contribute to improve the model robustness \wrt each other.

 \begin{table}[h]
 	\centering
 	\tabcolsep=0.1cm
 	\begin{tabular}{ |c||c|c|c|c|c|}
 		\hline
 		\multirow{2}{*}{Models Acc.(\%)}
 		& \multirow{2}{*}{Pristine}
 		& \multicolumn{4}{c|}{Test Time Attack} \\
 		\cline{3-6}
 		& &  \small{FGSM} & \small{PGD20} & \small{Spatial}  & \small{Joint}\\
 		\hline \hline
 		\alg{Natural} & {79.0} &  10.0 & 0.0 & 0.0 & 0.0  \\
 		\hline 
 		\alg{Madry} &  59.9 &  28.5 &  22.6  & 24.6  &  4.8  \\
 		\hline 
 		\alg{Bilateral} & 72.6  & 60.8  & 25.4  & 23.6   &  14.5  \\ 
 		\hline
 		\alg{Joint}  & 68.6 & {63.2} & {28.8} & {28.6} & {26.6} \\
 		\hline
 	\end{tabular}
 	\caption{Accuracy of different models under different attacks on CIFAR100 dataset ($\epsx\!=\!8, \epsw\!=\!0.3$).}
 	\label{tab:cifar100}
 \end{table}

\subsection{SVHN Dataset}
We further report results on the SVHN dataset~\cite{netzer2011reading}.
SVHN is a 10-way house number classification dataset, with  $73257$ training images and $26032$ test images. The additional training images are not used in experiment. The results are summarized in Table~\ref{tab:svhn}. The performance of the proposed approach compares favorably with  the models trained with~\cite{madry2017towards} and \cite{bilateral}.
More concretely, the proposed method outperforms all other methods under variants of pixel-based attacks (\emph{e.g.} FGSM, PGD), spatial attack as well as joint attack, indicating that the  integration of spatial with pixel attacks can not only improve model robustness against spatial attacks, which is lacking in the conventional models, but also boost the robustness against pixel attacks, for which the conventional models are specially designed for.
This further demonstrates the benefits of the proposed joint training approach.

\begin{table}[h]
	\centering
	\tabcolsep=0.1cm
	\begin{tabular}{ |c||c|c|c|c|c|}
		\hline
		\multirow{2}{*}{Models Acc.(\%)}
		& \multirow{2}{*}{Pristine}
		& \multicolumn{4}{c|}{Test Time  Attack} \\
		\cline{3-6}
		& &  \small{FGSM} & \small{PGD20} & \small{Spatial}  & \small{Joint}\\
		\hline \hline
		\alg{Natural} & 97.2 & 53.0  & 0.3 & 7.3 & 0.0 \\
		\hline 
		\alg{Madry} & 93.9 &	68.4 &	47.9&	76.7&	19.7 \\
		\hline 
		\alg{Bilateral} & 96.8 &	89.7&	55.1&	75.5&	35.0  \\ 
		\hline 
		\alg{Joint}  & {96.2} & {91.5} &	{56.4}&	{79.1}&	{38.8} \\
		\hline
	\end{tabular}
	\caption{Accuracy of different models under different attacks on SVHN dataset ($\epsx\!=\!8, \epsw\!=\!0.3$). }
	\label{tab:svhn}
	\vspace{-0.15in}
\end{table}

\section{Conclusion and Future Work}
We have presented a joint adversarial training approach in this paper.
Motivated by the goal of improving model robustness, we developed a spatial transformation-based attack method with explicit budget constraint, and presented an effective approach for joint adversarial attack generation and training incorporating both  spatial  and pixel attacks.
Extensive experiments on various datasets including CIFAR10, CIFAR100 and SVHN with comparison to state-of-the-art methods~\cite{madry2017towards,bilateral} verified the efficacy of the proposed method.
For future work, it is interesting to investigate how to further generalize the proposed approach to more general transformation, by leveraging more advanced differentiable renderers~\cite{OpenDR,halide,render_adv}.
While the current work and many other existing works  focused on small scale datasets, it would be interesting to  perform large scale joint adversarial training on   ImageNet~\cite{deng2009imagenet} following~\cite{kurakin2016scale,tramer2017ensemble,feature_dn}.

{\small
	\bibliographystyle{ieee}
	\bibliography{JAT}
}

\end{document}